\documentclass[letterpaper, 10 pt, conference]{ieeeconf}  

\IEEEoverridecommandlockouts                              

\usepackage{amsmath} 
\usepackage{amssymb}  
\usepackage{cite}
\usepackage{graphicx}
\usepackage{algorithm}
\usepackage{algpseudocode}
\usepackage{color}
\usepackage{float}
\usepackage{caption}

\title{\LARGE \textbf{SEAL}: \textbf{S}emantic Frame \textbf{E}xecution \textbf{A}nd \textbf{L}ocalization \\
for Perceiving Afforded Robot Actions
}

\author{Cameron Kisailus$^{1}$, Daksh Narang$^{1}$, Matthew Shannon$^{1}$ and Odest Chadwicke Jenkins$^{1}$
\thanks{$^{1}$C. Kisailus, D. Narang, and M. Shannon and O.C. Jenkins are with the Robotics Department,
        University of Michigan, Ann Arbor, MI, \{{\tt\small kisailus, dnarang, mattshan, ocj\}@umich.edu}}%
}

\usepackage{todonotes} 

\newcommand{\seal}{\textbf{S}emantic Frame \textbf{E}xecution \textbf{A}nd \textbf{L}ocalization for Perceiving Afforded Robot Actions }
\makeatletter
\let\@oldmaketitle\@maketitle
\renewcommand{\@maketitle}{\@oldmaketitle
  \includegraphics[width=\linewidth,height=20\baselineskip]
    {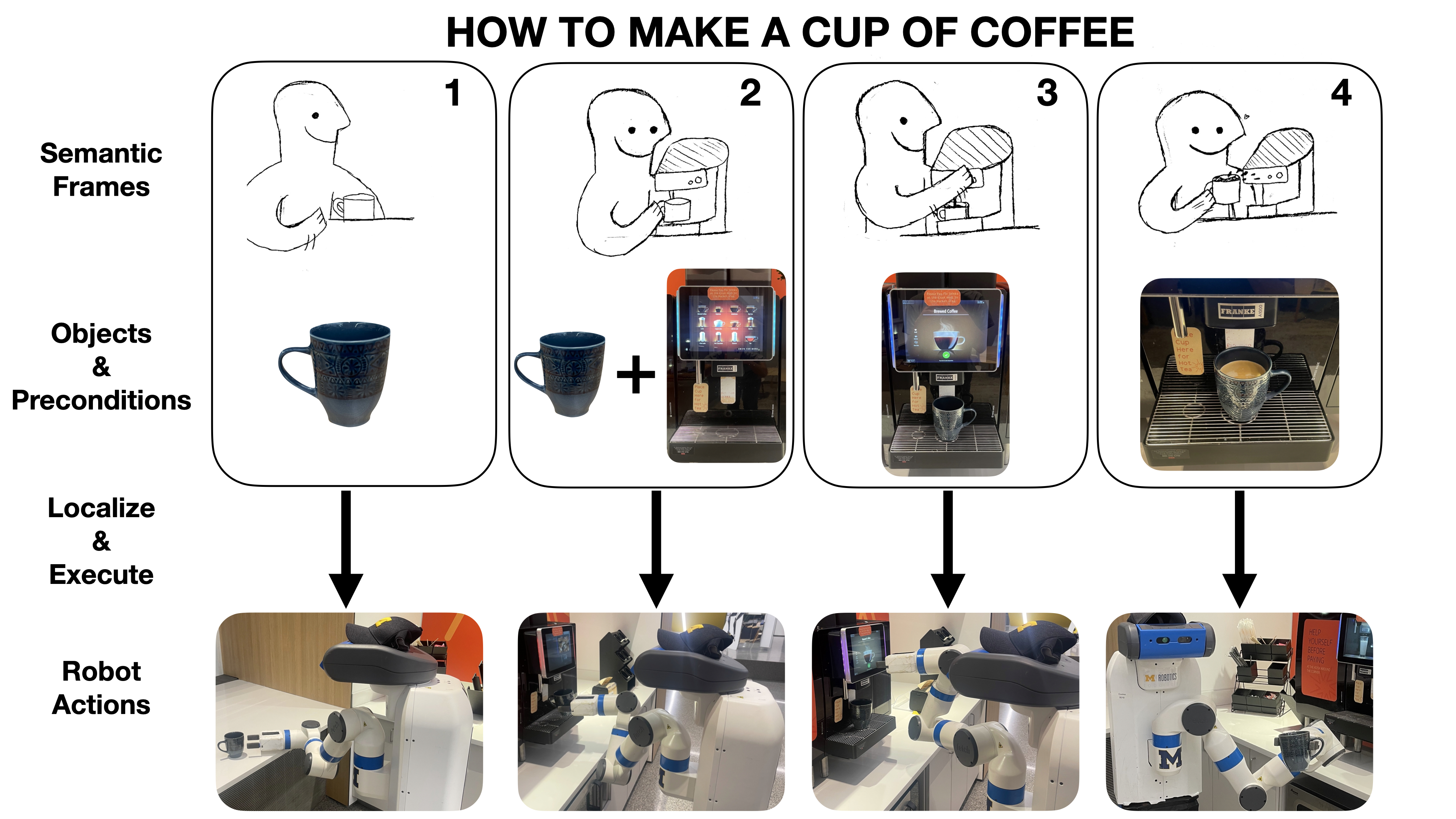}}
\makeatother

\begin{document}

\maketitle
\thispagestyle{empty}
\pagestyle{empty}

\begin{abstract}
Recent advances in robotic mobile manipulation have spurred the expansion of the operating environment for robots from constrained workspaces to large-scale, human environments. In order to effectively complete tasks in these spaces, robots must be able to perceive, reason, and execute over a diversity of affordances, well beyond simple pick-and-place. We posit the notion of semantic frames provides a compelling representation for robot actions that is amenable to action-focused perception, task-level reasoning, action-level execution, and integration with language.  Semantic frames, a product of the linguistics community, define the necessary elements, pre- and post- conditions, and a set of sequential robot actions necessary to successfully execute an action evoked by a verb phrase.  In this work, we extend the semantic frame representation for robot manipulation actions and introduce the problem of \seal(SEAL) as a graphical model. For the SEAL problem, we describe our nonparametric Semantic Frame Mapping (SeFM) algorithm for maintaining belief over a finite set of semantic frames as the locations of actions afforded to the robot. We show that language models such as GPT-3 are insufficient to address generalized task execution covered by the SEAL formulation and SeFM provides robots with efficient search strategies and long term memory needed when operating in building-scale environments.
\end{abstract}

\section{INTRODUCTION}
We envision autonomous systems that can perceive and perform tasks across large, building-scale spaces \cite{hawes2017}, \cite{veloso2015}, \cite{khandelwal2017} to serve needs across society, such as care-taking tasks in assisted living facilities and supply chain tasks in warehouses. In order to be effective, such systems must infer the objects present in the environment as well as predict the outcomes of actions afforded \cite{gibson1977} by these objects. In essence, robots need to perceive actions that are currently afforded by the environment, and not just the objects to be acted upon. For example, observing a cup should inform the system of an optimal location to achieve the action ``Grasp Cup”. In small enough workspaces, a robot can simply look for the objects required for the task, but as the environment grows this becomes infeasible. 

We are inspired by the idea that despite the aforementioned challenges, there is structure to human environments. Buildings, in most cases, are designed for efficient task completion by humans as objects and actions of similar types of usually in the vicinity of each other. For example, brooms, mops, and vacuums are likely to be in the closet whereas spoons, cups, and plates are likely to be in the kitchen. Moreover, we acknowledge the inherent structure of task execution due to the sequentiality of multi-step actions. While some affordances are inherent in certain object classes (i.e. \textit{Grasp} a cup, \textit{Open} a door, etc.), others have structured criteria, or preconditions, which must be met before being executed. For example, a cup must be full and near a container in order to \textit{Pour} the contents of the cup. Semantic frames, as elaborated in further sections, explicitly describe these relations and have been used in previous works to ground natural language commands in robot actions \cite{thomas2012}. Recently, the community has explored using Transformer-based models to ground natural language commands \cite{saycan} \cite{cliport} \cite{vemprala2023chatgpt}. While these models have shown impressive high-level reasoning capabilities, they often lack the physical intuition necessary to ground their output in feasible robot actions. 

Three core characteristics of semantic frames\cite{thomas2012} \cite{baker1998} \cite{ruppenhofer2016} motivate our exploration of their use as a representation to bring together language, action, and perception. First, they are evoked by a verb phrase such that we can directly parse natural language commands into semantic frames. Second, they explicitly define the objects necessary for execution. Last, they define the preconditions necessary before execution can begin and postconditions of the state after execution.  

In order to efficiently execute semantic frames, we require a model which can localize the frames location conditioned on observations of the environment.  Semantic perception of individual objects in large environments has been explored previously in the context of object search and generalized notions of object permanence \cite{zeng2020}.  We are now able to extend these ideas to consider  perception of afforded actions through inference over semantic frame representations.

In this paper, we introduce \seal (SEAL) which casts the affordance execution problem into a graphical model which accounts for object-affordance, state-affordance, and affordance-affordance relations. Additionally, we propose the Semantic Frame Mapping (SeFM) algorithm for perception of afforded actions in the context of task-level reasoning for mobile manipulation robot.  We consider SeFM as one possible algorithm for the broader SEAL problem. SeFM is a nonparametric particle-based inference method for maintaining belief over a finite set of semantic frames which represent the locations of actins afforded to the robot. We introduce and validate the SEAL model in a simulated apartment using ROS Gazebo. Next, we compare SeFM to a Transformer-based model on a multitude of household tasks using a simulated mobile manipulator and find that using SeFM leads to a higher success rate. Finally, we empower a real Fetch robot to execute tasks using SeFM.

\section{RELATED WORK}

\subsection{Generalizable Task Execution}
Generalized task execution has garnered much attention in the community as robotic perception and manipulation capabilities have improved. Recent works have demonstrated the ability to learn task-specific manipulation policies from RGB-D observations of the workspace \cite{shridhar2021} \cite{huang2022}. Though, in those works, they assume the environment is fully observable. Some attention has been given to operating in partially observable domains, but has met limited success due to challenges in perceiving the necessary objects for a task \cite{Inoue2023}. Other works have attempted to overcome the challenges imposed by human environments by utilizing a hybrid planning framework in which an online probabilistic semantic representation of the environment is passed to an offline task planner \cite{wang2022}, or by restricting the action space to unstructured (i.e. atomic) actions that have a uniform likelihood throughout the environment \cite{sarch2022tidee}. Recent methods have explored the use of Large Language Models (LLM) as planners \cite{saycan}, \cite{vemprala2023chatgpt}. While LLM do show promising results in reasoning over high-level goals, they struggle to ground their output in robot actions, even with appropriate prompting. Additionally, LLM cannot inherently estimate whether an action is afforded in the current scene, so in \cite{saycan} the authors train value function offline mapping RGB images of the state to executability and \cite{vemprala2023chatgpt} they restrict the output to API function names which are always actionable. Our proposed method, SEAL, addresses these challenges by formulating affordance execution as a search problem. Estimating executability is no long necessary as we can simply cross-reference the defined preconditions with the known state of the world. Partial observability is overcome through actively searching the environment for necessary frame elements. 

\subsection{Semantic Frames}
Semantic frames \cite{thomas2012},\cite{ruppenhofer2016} describe affordances, complete with actors, objects, preconditions, and results. A semantic frame is said to be evoked by a particular verb clause making them good representations of actions due to their implicit ability to generalize task description across variations in environment, object instances, and even request phrases. “Get Roger a coffee” and “Bring Roger a coffee” evoke the same semantic frame: ``Bring \{\textit{object}\} to \{\textit{recipient}\}". Formally, a semantic frame, $f$,  is defined as $f = (O, P, A)$,  where $O$, $P$, and $A$ are the sets of frame elements (objects), preconditions, and robot actions encoded in the frame, respectively. Semantic frames also have a notion of postconditions --- how the state transitions given a successful frame execution --- which can be logically sequenced to generate task plans for high level goals. Previous work \cite{thomas2012} \cite{ruppenhofer2016} has shown the ability to ground natural language commands into robot actions by parsing commands into semantic frames. In \cite{thomas2012}, $A$ consisted only of locomotion actions. In this work, we expand the set of possible actions to include manipulation and move toward performing complex tasks across large building-wide spaces. Because it is now necessary to interact with objects, a new problem of semantic frame localization is introduced. 

\subsection{Conditional Random Fields}
Conditional Random Fields (CRF) \cite{Sutton2012} \cite{lafferty2001} are a class of statistical modeling methods introduced in machine learning for sequence labelling problems. In essence, CRFs are an extension of Hidden Markov Models \cite{Rabiner1986} with the ability to incorporate complex, higher-order dependencies among the input features. CRFs are particularly useful when the outputs are correlated and depend not only on the current input but also on the context of neighboring inputs. A factor graph is a probabilistic graphical model where the nodes represent variables and edges represent the conditional dependencies between variables. In the case of CRFs, the variables are inputs and outputs, and the edges are the dependencies neighboring input features and outputs. Prior work \cite{zeng2020} has shown that casting the object search problem into a CRF factor graph can lead to performance gains in partially observable, real-world environments without the need for strong assumptions about static landmarks as in \cite{kollar2009} \cite{kunze2014} \cite{toris2017}.

\begin{figure}
    \centering
    \includegraphics[height=6cm]{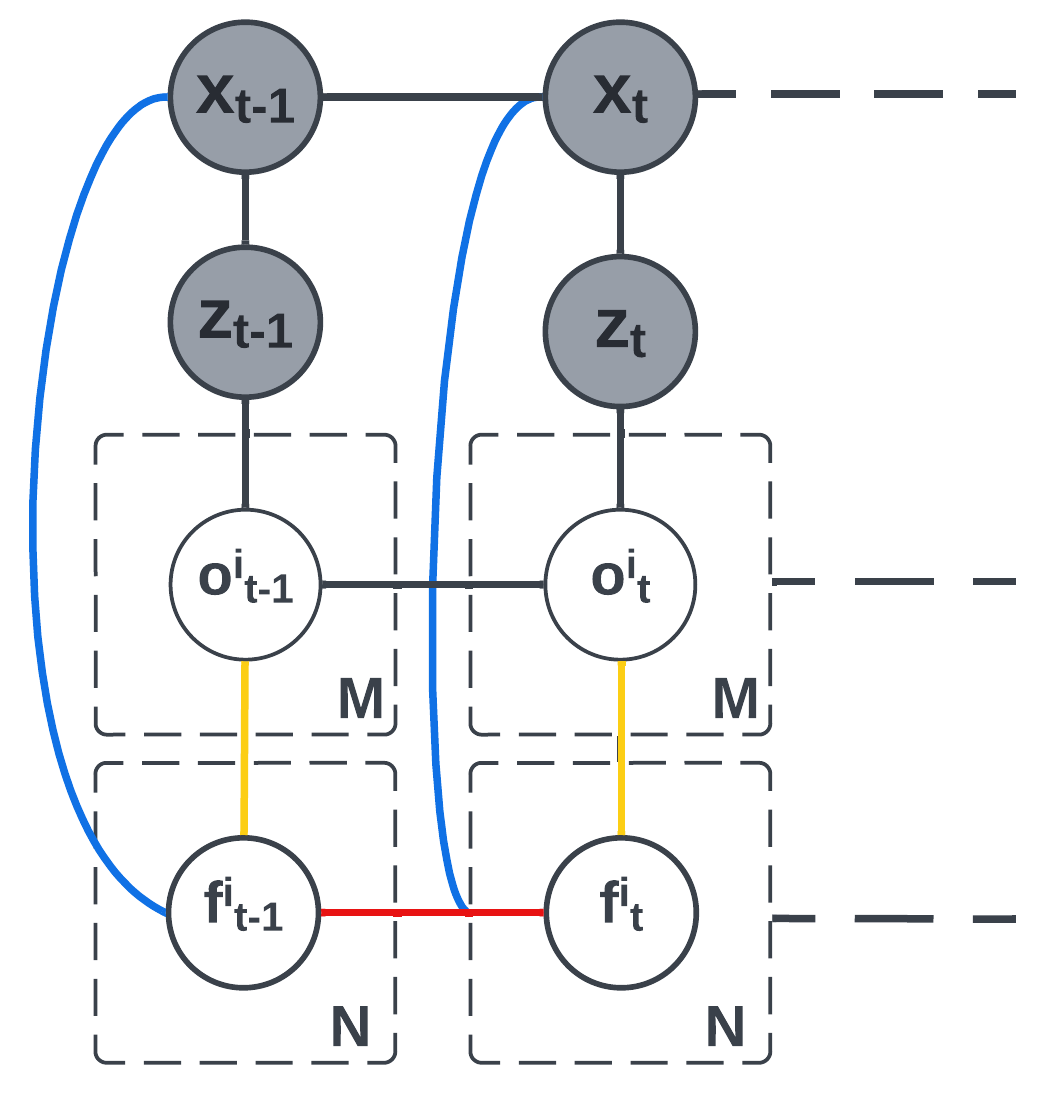}
    \caption{SEAL Model: Known: ${x_t}$ robot-state information, $z_t$ sensor observations; Unknown: $O_t:\{o^i:i \in 1:M\}$ object locations, $F_t:\{f^i:i \in 1:N\}$ frame locations. Prediction potential shown in red, measurement potential shown in yellow, context potential shown in blue. }
    \label{fig:CRF}
\end{figure}
\section{Method}
\subsection{Problem Formulation}
Let $F=\{f^i | i=1,..,N\}$ be the set of semantic frames we wish to localize. Let $O=\{o^i|i=1,...,M\}$ be the collective set of all objects defined in the frame elements of all $f \in F$. Given observations $z_{0:T}$ and robot-state information $x_{0:T}$, we wish to maintain the belief over frame locations $P(F_{0:T} | x_{0:T}, z_{0:T}, O_{0:T})$. The resulting distribution will inform the robot where affordances are in the scene, and, thus, where the robot can take actions and effect change in the environment. We assume a room-level annotated metric map is given. SEAL describes the inference problem of localizing instances of semantic frames within the given map.

\subsection{SEAL} \label{sec:SeFM}
SEAL formalizes the frame location estimation problem via a Conditional Random Field (CRF) extending the work of Lorbach \cite{lorbach2014} and Zeng \cite{zeng2020}.  The CRF model includes object-affordance, affordance-affordance, and state-affordance relations as shown in Figure \ref{fig:CRF}. The full posterior probability of frame locations is

\begin{multline}
        p(F_{0:T}|x_{0:T}, z_{0:T}, O_{0:T}) =\\\frac{1}{Z} \prod_{t=0}^{T}\prod_{i=1}^{N}\phi_p(f^i_t, f^i_{t-1})\prod_{i,k}\phi_{m,\mathcal{B}(R_{ik}|x_t)}(f^i_t, o^k_t)\\\prod_{i,j}\phi_{c,\mathcal{B}(R_{ij}|x_t)}(f^i_t, f^j_t)
    \label{eq:full_posterior}
\end{multline}

where Z is a normalization constant, $\phi_p$ is the prediction potential, $\phi_m$ is the measurement potential, and $\phi_c$ is the context potential. We assume the robot remains localized in a metric map of the environment. Robot-state information $x_t$ informs the model about concepts regarding the robot itself: pose of the robot, whether an object is in its gripper, etc. Observations $z_t$ are RGB-D images of the environment taken by the robot as it navigates. Both $x_t$ and $z_t$ are known variables. 

The \textit{prediction potential}, $\phi_p(f^i_t, f^i_{t-1})$, models the temporal permanence of a frame. The value of this potential is category-dependent. Some frames, like ``\textit{Go} to couch", remain static over time. Others, like ``\textit{Grasp} cup", can move over time due to various external factors which we model as:

\begin{equation*}
    \phi_p(f^i_t, f^i_{t-1}) \sim \mathcal{N}(f^i_{t-1}, \Sigma^i)
    \label{eq:prediction}
\end{equation*}
The \textit{measurement potential} ,$\phi_{m,\mathcal{B}(R_{ik}|x_t)}(f^i_t, o^k_t)$, models object-frame relations parameterized by the belief --- $\mathcal{B}(R_{ik}|x_t)$ --- over a set of defined frame-object relations $R$ for frame $f^i$ and object $o^k$ conditioned on robot-state information $x_t$. We assume that frames are tightly coupled to the objects which afford them (i.e. ``\textit{Grasp} Spoon" is spatially close to a spoon). By parameterizing the belief over all frame elements, we are able to model the effect of state-transition on frame locations. For example, ``\textit{Stir} Mug" requires a spoon and the semantic frame explicitly encodes this with the precondition ``\textit{Grasp} spoon".  Therefore, if the robot does not yet have a spoon in its gripper ``\textit{Stir} Mug" should be localized close to a spoon; conversely, ``\textit{Stir} Mug" should be close to a mug if the robot is already holding a spoon. Additionally, because we model the relation over all frame elements ``\textit{Stir} Mug" is more likely to be in a room that has both spoons and mugs rather than a room that only has one or the other. Concretely, 
\begin{equation*}
    \phi_{m,\mathcal{B}(R_{ik}|x_t)} = \sum_r\sum_{k=0}^M \mathcal{B}(R_{ik} = r | x_t) \phi_{m,r}(f^i,o^k, R_{ik}=r)
    \label{eq:measurement}
\end{equation*}
where $r$ can be one of the following relation types \{\textit{Core}, \textit{Other}, \textit{Disjoint}\} and $\mathcal{B}(R_{ik} = r | x_t)$ is the belief that $r$ is relevant between a frame ($f^i$) and object ($o^k$) given the current state. For $r \in $ \{\textit{Core}, \textit{Other}\}, the measurement potential corresponds to a Gaussian distribution:
\begin{equation*}
    \phi_{m}(f_t^i, o_t^k) \sim \mathcal{N}(o^k_t, \Sigma)
\end{equation*}
where $\Sigma$ is always constant across frame-object pairs. A \textit{Core} object is the next object the robot would need to interact with to proceed with frame execution. Whereas \textit{Other} objects will eventually be necessary or are completely optional.
For $r = $ \textit{Disjoint} the measurement potential is 0 since the object is not involved in frame execution. In this work, object-frame relations are explicit in the semantic frame definition.

The \textit{context potential} ,$\phi_{c,\mathcal{B}(R_{ij})}(f^i_t, f^j_t, x_t)$, models inter-frame relations. In this work we only model the relation between a frame and its preconditions. Since the preconditions define a sequential list of actions to be completed, we can use the state information to inform our model about which preconditions have already been met and which precondition must be met next. From there, the context potential follows a similar model to the measurement potential 
\begin{equation*}
    \phi_{c,\mathcal{B}(R_{ij}|x_t)} = \sum_r\sum_{j=0}^M \mathcal{B}(R_{ij} = r | x_t) \phi_{c,r}(f^i,f^j, R_{ij}=r)
    \label{eq:context}
\end{equation*}
where $r$ can be one of two relation types \{\textit{Precondition}, \textit{Disjoint}\}. For $r = $ \textit{Precondition} the context potential corresponds to a Gaussian distribution:
\begin{equation*}
    \phi_{c}(f_t^i, f_t^j) \sim \mathcal{N}(f^j_t, \Sigma)
\end{equation*}
where $f^j$ is the precondition and $\Sigma$ is always constant. For $r = $ \textit{Disjoint} the context potential is 0. Again, these relations are explicit in the semantic frame definition.

\subsection{Semantic Frame Mapping} \label{sec:inference}
We implement a particle-based inference method, dubbed Semantic Frame Mapping, for maintaining belief over semantic frame locations, as shown in Algorithm \ref{algo:inference}. Object locations are inferred using the method defined in \cite{zeng2020} and we refer the reader to that paper for full implementation details. In our experiments, we use 200 particles for each object and semantic frame with heurisitics for particle reinvigoration. 

\begin{algorithm}[!th]
    \caption{Inference of Semantic Frame Locations in SeFM}
    \label{algo:inference}
    \begin{algorithmic}
        \Require Observation $z_t$, Robot-State Vector $x_t$, Object location particle set $O_{t}$, particle set for each frame:\\
        $f^i_{t-1}=\{\langle f^{i(k)}_{t-1}, \alpha^{i(k)}_{t-1} \rangle | k=1,...,P \}, i \in 1:N$
        \State Resample P particles with probability proportional to $\alpha_{t-1}^{i(k)}$
        \For{$i=1,...,N$}
            \For{$k=1,...,P$}
                \State $f^{i(k)}_t \sim \phi_p(f^i_t, f^{i(k)}_{t-1})$
                \State $\displaystyle \alpha_t^{i(k)} \propto \! \prod_{j \in \Gamma(i)} \! \phi_{m}(f^{i(k)}_t, o^j_t) \prod_{l \in \Gamma(i)}  \! \phi_{c} (f^{i(k)}_t, f^l_t)$
                
                where, 
                $\displaystyle \phi_{m}= \sum_r\sum^P_{s=0}\mathcal{B}(r|x_t)\alpha_{t}^{j(s)}\phi_{m,r}(f^i_t, o^j_t,r) $
                
                and,
                $\displaystyle \phi_{c}= \sum_r\sum^P_{s=0}\mathcal{B}(r|x_t)\alpha_{t}^{l(s)}\phi_{c,r}(f^i_t, f^l_t,r)$
        \EndFor
    \EndFor
    \end{algorithmic}
\end{algorithm}

\section{RESULTS}
\begin{figure}
    \centering
    \includegraphics[width=\columnwidth]{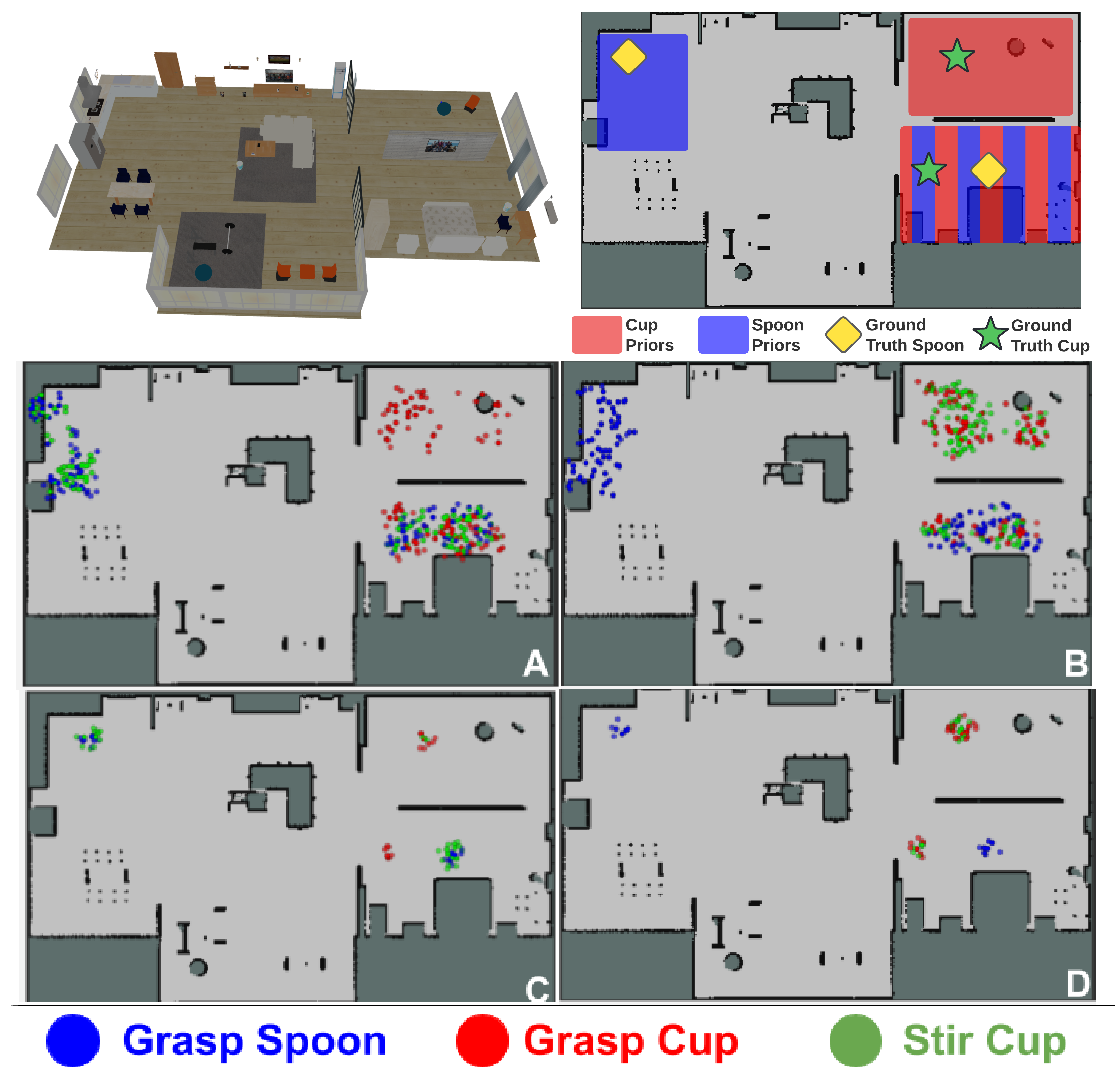}
    \caption{SeFM Inference in Gazebo Apartment Setting. Object's room-level priors (Red=Cup, Blue=Spoon) and ground truth locations (Diamond=Spoon, Star=Cup) shown in top right. Resulting particle distributions shown in A,B,C,D. A\&B validate context potential as Stir Cup density (Green) shifts from close to spoon (A) to cup (B) depending on whether robot is currently grasping a spoon. C\&D validate measurement potential as distributions converge when objects are observed.}
    \label{fig:gazebo_results}
\end{figure}
\subsection{Inference}
\begin{figure*}
    \includegraphics[width=\linewidth]{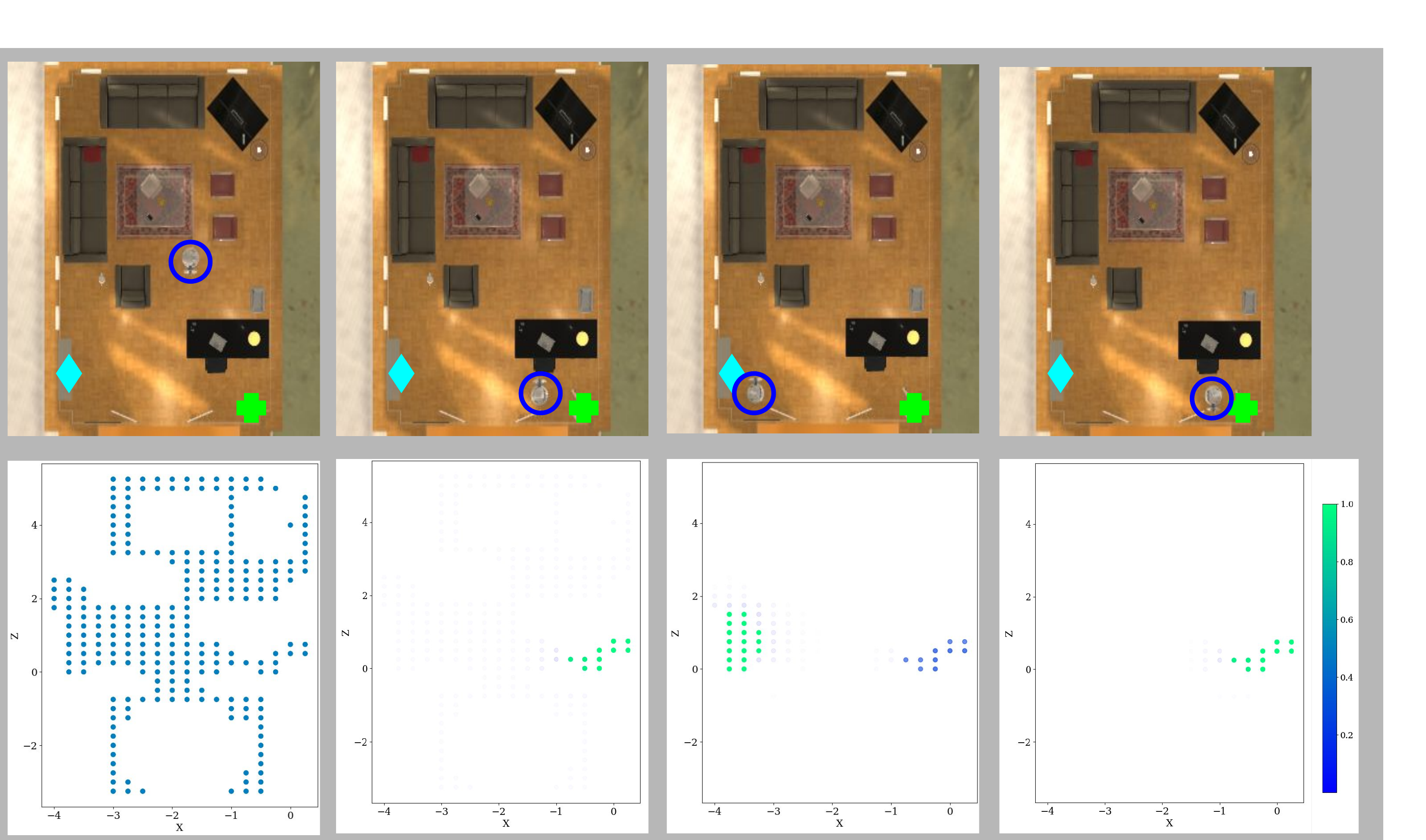}
    \caption{SeFM implemented in iTHOR simulation tasked with ``Put Vase in Safe". Top row shows topdown view of environment with Robot circled in blue, Vase location marked with teal diamond, and Safe marked with green cross. Bottom row shows the distribution at various timesteps throughout the episode. Beginning with initial, uniform distribution, particle weights are updated according to Algorithm \ref{algo:inference}.}
    \label{fig:iTHOR}
\end{figure*}
\label{sec:gazebo_inf}
We begin by studying the effectiveness of SEAL to model the location of affordances in a simulated apartment environment using ROS Gazebo. Figure \ref{fig:gazebo_results} shows our experimental setup. The shaded rectangles represent room-level priors explicitly given to the agent a priori. The agent does not know the ground truth \textit{Spoon} and \textit{Cup} locations a priori, but can observe them while exploring the environment. Particles are uniformly initialized throughout the map for each frame and object class. We choose to maintain belief over the semantic frames ``\textit{Grasp} Spoon", ``\textit{Grasp} Cup" and ``\textit{Stir} Cup". Where ``\textit{Stir} Cup" refers to the action of grasping a spoon then using that spoon to stir the contents of a cup.

To explore SEAL's effectiveness under partial observability, we keep the agent at a fixed position where none of the objects are observable. Figure \ref{fig:gazebo_results} A-B show the resulting particle distributions after 20 belief update iterations. In Figure \ref{fig:gazebo_results}A, the agent is initialized with empty grippers whereas in \ref{fig:gazebo_results}B the agent is initialized already grasping a spoon. The effect of this change is reflected in the final distribution of ``\textit{Stir} Cup". Without a spoon (\ref{fig:gazebo_results}A), we maintain belief near likely locations of a spoon. Furthermore, since we sum over all frame elements, we maintain relatively higher density in the region where both a spoon and cup are likely to be. When the agent has a spoon (\ref{fig:gazebo_results}B), the density of ``\textit{Stir} Cup" shifts to locations where cups are likely to be, since the measurement potential between spoon and ``\textit{Stir} Cup" is now 0. This result affirms that SEAL can accurately condition semantic frame locations $f_t$ on the robot-state information vector $x_t$ through the context potential $\phi_{c,\mathcal{B}(R_{ij})}(f^i_t, f^j_t, x_t)$. 

To incorporate observations from the environment, we have the agent follow a predefined trajectory through the environment. We assume the agent has a 5m observable range. Figures \ref{fig:gazebo_results} C-D display the particle distributions after the agent has finished navigation. Here we follow the same initialization routine as mentioned above (i.e. C is initialized without a spoon and D is). By incorporating observations, the final distributions converge to observed objects and show the same sensitivity to initial conditions as in \ref{fig:gazebo_results}A-B. This suggests that the measurement potential $\phi_{m,\mathcal{B}(R_{ik}|x_t)}$ is effective in the convergence of frame locations when frame elements are observed. We later incorporate this into an active search algorithm in simulated and real robots.

\subsection{Task Execution}
\label{sec:iTHOR}
Now we explore the utility these distributions are when a robot is tasked with executing a semantic frame. Experiments are conducted using a mobile manipulator in the iTHOR simulation environment \cite{ai2thor} using tasks from the ALFRED benchmark \cite{ALFRED20}. ALFRED is a public benchmark used to evaluate the ability to ground natural language commands for everyday household tasks. Tasks in ALFRED are commanded using a natural language sentence, domains are kitchens, bathrooms, and living rooms and contain actions with irreversible state changes. To apply SEAL to this, we slightly alter our inference method to now reason over robot poses that allow for interaction with an affordance rather than the affordance location itself. Additionally, in this case, we no longer use room-level priors for objects since the operating domain is single-room. When given a task, we first parse the task into a set of semantic frames. Next, the agent uses our particle-based inference method over SEAL to infer the poses at which a semantic frame can be executed, and, finally, navigates to and executes the action primitive defined in the semantic frame. Figure \ref{fig:iTHOR} shows the progression of distributions for the semantic frame ``Put Vase in Safe".

We choose a subset of 50 trials from each of the following experiment groups in ALFRED: Look at, Pick-Place, Pick-Stack-Place, and Pick-Heat-Place and refer the reader to the original work for task descriptions. We define 6 semantic frames (Pick, Place, Slice, Open, Close and Heat) grounded in singular action events that can be called in the iTHOR simulator. We compare SeFM to a method similar to SayCan \cite{saycan} in which a Large Language Model is queried to provide the next robot action conditioned on current state and commanded goal. In this experiment, we query GPT-3 using OpenAI's API client. Further, we use the same affordance scoring algorithm available in the public SayCan implementation based on object detection rather then a learned value function. 

\begin{figure}
    \label{fig:results}
    \includegraphics[width=\columnwidth]{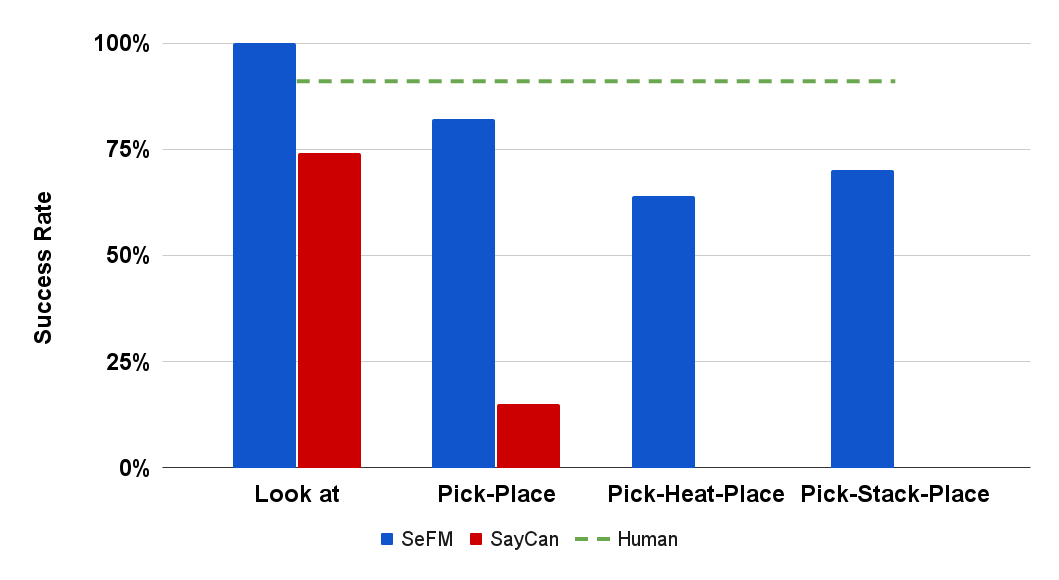}
    \caption{Success rate of SeFM (Blue) and SayCan (Red) across each task group. Human performance shown as green dashed line.} 
\end{figure}
Figure 4 shows the success rate of each algorithm across the 4 aforementioned task groups. Success rate here is defined as the percentage of trials which completed all the required actions in the correct order; partial completion of a task counts as a failure. We note that SeFM does significantly better than SayCan, especially in multi-step tasks. Empirically, we found that GPT-3 will often propose actions which are not yet afforded to the robot. For example, when heating an object GPT-3 will suggest ``Turn on Microwave" prior to suggesting ``Close Microwave" leading to failure. Because semantic frames explicitly encode these preconditions, SeFM does not struggle with this. Further, we found that Saycan does poorly when a required object is not immediately observable by the robot. This could be alleviated by improving the affordance scoring function, but that requires additional training data. As shown in section \ref{sec:gazebo_inf}, SeFM maintains an informed belief of affordance locations even without observing the necessary objects. This ability allows for the robot to search its environment efficiently for necessary objects, boosting performance here.

\subsection{Real Robot}
Finally, we implement SeFM on a Fetch mobile manipulator. We start by creating a 2D occupancy map of the operating environment and annotate said map with known regions (i.e. "Lab", "Hallway", "Kitchenette", etc.). Robot-state information, $x_t$, maintains knowledge of the pose of the robot in the map frame, a history of semantic frames the robot has previously executed, and the name of the object currently in the gripper. For our observations, we use a pretrained YOLOv7 \cite{wang2022yolov7} network finetuned on real world images of objects from the YCB object dataset \cite{Calli2015}. To determine navigation goals, we use a similar method to \cite{zeng2020} which fits a Bayesian Gaussian Mixture Model to the distribution and chooses a pose which allows the robot to observe the mean of the resulting Gaussian. Action policies (Pick, Place, etc.) are manually engineered using MoveIt!. We task the robot with tasks similar to those described in Section \ref{sec:iTHOR} -- excluding Pick-Heat-Place due to Fetch's inability to operate a microwave. 10 trials are conducted for each experiment group. We achieve success rates of 80\%, 60\% and 20\% for Look at, Pick-Place, and Pick-Stack-Place, respectively. We note that a majority of failures came from errors during manipulation not inference or navigation. 
\section{CONCLUSION}
We show that Semantic Frame Mapping can incorporate the structure of human environments as factors in a factor graph to efficiently and accurately localize semantic frames even with weak priors and under partial observability. We also show that by using particle based inference methods we are able to encode multi-modal belief distributions efficiently and without mode collapse. We explored the advantages of Semantic Frame decomposition of tasks compared to LLM and showed that LLM fail to grasp the subtleties of complex actions even when prompted with an example. This shortcoming led to 0\% success rate in tasks that SeFM was able to complete over 50\% of the time. When extending SeFM to the real world, we noticed that a majority of failures come from the encoded manipulation policies failing. With more refinement in this domain, we expect Semantic Frame Mapping to be a viable bridge from high-level command to robot actions. Further work remains to remove the necessity to explicitly define each component of a semantic frame; for instance, could we learn the frame elements, pre- and post- conditions from demonstrations? 
\bibliographystyle{ieeetr}
\bibliography{biblio}

\begin{thebibliography}{10}

\bibitem{hawes2017}
N.~Hawes, C.~Burbridge, F.~Jovan, L.~Kunze, B.~Lacerda, L.~Mudrova, J.~Young,
  J.~Wyatt, D.~Hebesberger, T.~Kortner, R.~Ambrus, N.~Bore, J.~Folkesson,
  P.~Jensfelt, L.~Beyer, A.~Hermans, B.~Leibe, A.~Aldoma, T.~Faulhammer,
  M.~Zillich, M.~Vincze, E.~Chinellato, M.~Al-Omari, P.~Duckworth,
  Y.~Gatsoulis, D.~C. Hogg, A.~G. Cohn, C.~Dondrup, J.~Pulido~Fentanes,
  T.~Krajnik, J.~M. Santos, T.~Duckett, and M.~Hanheide, ``The strands project:
  Long-term autonomy in everyday environments,'' {\em IEEE Robotics \&
  Automation Magazine}, vol.~24, no.~3, pp.~146--156, 2017.

\bibitem{veloso2015}
M.~Veloso, J.~Biswas, B.~Coltin, and S.~Rosenthal, ``Cobots: Robust symbiotic
  autonomous mobile service robots,'' in {\em Twenty-Fourth International Joint
  Conference on Artificial Intelligence}, 2015.

\bibitem{khandelwal2017}
P.~Khandelwal, S.~Zhang, J.~Sinapov, M.~Leonetti, J.~Thomason, F.~Yang,
  I.~Gori, M.~Svetlik, P.~Khante, V.~Lifschitz, {\em et~al.}, ``Bwibots: A
  platform for bridging the gap between ai and human--robot interaction
  research,'' {\em The International Journal of Robotics Research}, vol.~36,
  no.~5-7, pp.~635--659, 2017.

\bibitem{gibson1977}
J.~J. Gibson, ``The theory of affordances,'' {\em Hilldale, USA}, vol.~1,
  no.~2, pp.~67--82, 1977.

\bibitem{thomas2012}
B.~J. Thomas and O.~C. Jenkins, ``Roboframenet: Verb-centric semantics for
  actions in robot middleware,'' in {\em 2012 IEEE International Conference on
  Robotics and Automation}, pp.~4750--4755, IEEE, 2012.

\bibitem{saycan}
M.~Ahn, A.~Brohan, N.~Brown, Y.~Chebotar, O.~Cortes, B.~David, C.~Finn, C.~Fu,
  K.~Gopalakrishnan, K.~Hausman, A.~Herzog, D.~Ho, J.~Hsu, J.~Ibarz, B.~Ichter,
  A.~Irpan, E.~Jang, R.~J. Ruano, K.~Jeffrey, S.~Jesmonth, N.~J. Joshi,
  R.~Julian, D.~Kalashnikov, Y.~Kuang, K.-H. Lee, S.~Levine, Y.~Lu, L.~Luu,
  C.~Parada, P.~Pastor, J.~Quiambao, K.~Rao, J.~Rettinghouse, D.~Reyes,
  P.~Sermanet, N.~Sievers, C.~Tan, A.~Toshev, V.~Vanhoucke, F.~Xia, T.~Xiao,
  P.~Xu, S.~Xu, M.~Yan, and A.~Zeng, ``Do as i can, not as i say: Grounding
  language in robotic affordances,'' 2022.

\bibitem{cliport}
M.~Shridhar, L.~Manuelli, and D.~Fox, ``Cliport: What and where pathways for
  robotic manipulation,'' in {\em Proceedings of the 5th Conference on Robot
  Learning (CoRL)}, 2021.

\bibitem{vemprala2023chatgpt}
S.~Vemprala, R.~Bonatti, A.~Bucker, and A.~Kapoor, ``Chatgpt for robotics:
  Design principles and model abilities,'' Tech. Rep. MSR-TR-2023-8, Microsoft,
  February 2023.

\bibitem{baker1998}
C.~F. Baker, C.~J. Fillmore, and J.~B. Lowe, ``The berkeley framenet project,''
  in {\em COLING 1998 Volume 1: The 17th International Conference on
  Computational Linguistics}, 1998.

\bibitem{ruppenhofer2016}
J.~Ruppenhofer, M.~Ellsworth, M.~Schwarzer-Petruck, C.~R. Johnson, and
  J.~Scheffczyk, ``Framenet ii: Extended theory and practice,'' tech. rep.,
  International Computer Science Institute, 2016.

\bibitem{zeng2020}
Z.~Zeng, A.~R{\"o}fer, and O.~C. Jenkins, ``Semantic linking maps for active
  visual object search,'' in {\em 2020 IEEE International Conference on
  Robotics and Automation (ICRA)}, pp.~1984--1990, IEEE, 2020.

\bibitem{shridhar2021}
M.~Shridhar, L.~Manuelli, and D.~Fox, ``Cliport: What and where pathways for
  robotic manipulation,'' in {\em Proceedings of the 5th Conference on Robot
  Learning (CoRL)}, 2021.

\bibitem{huang2022}
H.~Huang, D.~Wang, R.~Walter, and R.~Platt, ``Equivariant transporter
  network,'' {\em arXiv preprint arXiv:2202.09400}, 2022.

\bibitem{Inoue2023}
Y.~Inoue and H.~Ohashi, ``Prompter: Utilizing large language model prompting
  for a data efficient embodied instruction following,'' 2022.

\bibitem{wang2022}
Z.~Wang and G.~Tian, ``Hybrid offline and online task planning for service
  robot using object-level semantic map and probabilistic inference,'' {\em
  Information Sciences}, vol.~593, pp.~78--98, 2022.

\bibitem{sarch2022tidee}
G.~Sarch, Z.~Fang, A.~W. Harley, P.~Schydlo, M.~J. Tarr, S.~Gupta, and
  K.~Fragkiadaki, ``Tidee: Tidying up novel rooms using visuo-semantic
  commonsense priors,'' {\em arXiv preprint arXiv:2207.10761}, 2022.

\bibitem{Sutton2012}
C.~Sutton and A.~McCallum, ``An introduction to conditional random fields,''
  {\em Foundations and Trends® in Machine Learning}, vol.~4, no.~4,
  pp.~267--373, 2012.

\bibitem{lafferty2001}
J.~Lafferty, A.~McCallum, and F.~C. Pereira, ``Conditional random fields:
  Probabilistic models for segmenting and labeling sequence data,'' 2001.

\bibitem{Rabiner1986}
L.~Rabiner and B.~Juang, ``An introduction to hidden markov models,'' {\em IEEE
  ASSP Magazine}, vol.~3, no.~1, pp.~4--16, 1986.

\bibitem{kollar2009}
T.~Kollar and N.~Roy, ``Utilizing object-object and object-scene context when
  planning to find things,'' in {\em 2009 IEEE International Conference on
  Robotics and Automation}, pp.~2168--2173, IEEE, 2009.

\bibitem{kunze2014}
L.~Kunze, K.~K. Doreswamy, and N.~Hawes, ``Using qualitative spatial relations
  for indirect object search,'' in {\em 2014 IEEE international conference on
  robotics and automation (ICRA)}, pp.~163--168, IEEE, 2014.

\bibitem{toris2017}
R.~Toris and S.~Chernova, ``Temporal persistence modeling for object search,''
  in {\em 2017 IEEE international conference on robotics and automation
  (ICRA)}, pp.~3215--3222, IEEE, 2017.

\bibitem{lorbach2014}
M.~Lorbach, S.~H{\"o}fer, and O.~Brock, ``Prior-assisted propagation of spatial
  information for object search,'' in {\em 2014 IEEE/RSJ International
  Conference on Intelligent Robots and Systems}, pp.~2904--2909, IEEE, 2014.

\bibitem{ai2thor}
E.~Kolve, R.~Mottaghi, W.~Han, E.~VanderBilt, L.~Weihs, A.~Herrasti, D.~Gordon,
  Y.~Zhu, A.~Gupta, and A.~Farhadi, ``{AI2-THOR: An Interactive 3D Environment
  for Visual AI},'' {\em arXiv}, 2017.

\bibitem{ALFRED20}
M.~Shridhar, J.~Thomason, D.~Gordon, Y.~Bisk, W.~Han, R.~Mottaghi,
  L.~Zettlemoyer, and D.~Fox, ``{ALFRED: A Benchmark for Interpreting Grounded
  Instructions for Everyday Tasks},'' in {\em The IEEE Conference on Computer
  Vision and Pattern Recognition (CVPR)}, 2020.

\bibitem{wang2022yolov7}
C.-Y. Wang, A.~Bochkovskiy, and H.-Y.~M. Liao, ``{YOLOv7}: Trainable
  bag-of-freebies sets new state-of-the-art for real-time object detectors,''
  {\em arXiv preprint arXiv:2207.02696}, 2022.

\bibitem{Calli2015}
B.~Calli, A.~Walsman, A.~Singh, S.~Srinivasa, P.~Abbeel, and A.~M. Dollar,
  ``Benchmarking in manipulation research: Using the yale-{CMU}-berkeley object
  and model set,'' {\em {IEEE} Robotics {\&}amp Automation Magazine}, vol.~22,
  pp.~36--52, sep 2015.

\end{thebibliography}
\end{document}